\documentclass[12pt, reqno]{amsart}
\usepackage{amsmath, amsthm, amscd, amsfonts, amssymb, graphicx, xcolor}
\usepackage[bookmarksnumbered, colorlinks, plainpages]{hyperref}

\theoremstyle{definition}

\theoremstyle{remark}

\numberwithin{equation}{section}

\usepackage{graphicx}

\begin{document}
\setcounter{page}{1}

\centerline{}

\centerline{}


\title[Cinder]{Cinder: A fast and fair matchmaking system}

\author[S. Pal]{Saurav Pal$^1$$^{*}$}

\address{$^{1}$ Department of Computer Science and Engineering, National Institute of Technology, Silchar, India.}
\email{\textcolor[rgb]{0.00,0.00,0.84}{palsaurav.2020@gmail.com}}


\date{Published: November 2, 2025.
\newline \indent $^{*}$ Corresponding author
\newline \indent © Saurav Pal 2025. This article is licensed under a Creative Commons Attribution 4.0
International License. To view a copy of the licence, visit 
\newline \indent \url{https://creativecommons.org/licenses/by/4.0/}}

\begin{abstract}
A fair and fast matchmaking system is an important component of modern multiplayer online games, directly impacting player retention and satisfaction. However, creating fair matches between lobbies (pre-made teams) of heterogeneous skill levels presents a significant challenge. Matching based simply on average team skill metrics, such as mean or median rating or rank, often results in unbalanced and one-sided games, particularly when skill distributions are wide or skewed. This paper introduces Cinder, a two-stage matchmaking system designed to provide fast and fair matches. Cinder first employs a rapid preliminary filter by comparing the ``non-outlier" skill range of lobbies using the Ruzicka similarity index. Lobbies that pass this initial check are then evaluated using a more precise fairness metric. This second stage involves mapping player ranks to a non-linear set of skill buckets, generated from an inverted normal distribution, to provide higher granularity at average skill levels. The fairness of a potential match is then quantified using the Kantorovich distance \cite{kantorovich, vaserstein} on the lobbies' sorted bucket indices, producing a ``Sanction Score." We demonstrate the system's viability by analyzing the distribution of Sanction Scores from 140 million simulated lobby pairings, providing a robust foundation for fair matchmaking thresholds.
\end{abstract} \maketitle


\section{Introduction}
\label{sec:introduction}

In the competitive online gaming scene, the matchmaking system serves as the foundational mediator of player experience. Its primary directive is to pair waiting teams for a match that is both fair and engaging. This task is trivial when all players in the queue are of similar skill, or ``rating" or ``rank" (as illustrated in Fig. \ref{fig:equal-rank}). However, modern games actively encourage social play, allowing players to form ``lobbies" or pre-made parties with friends who may have widely divergent skill levels.

This social dynamic introduces significant complexity. A single lobby may have a wide rank distribution, be evenly spread across the skill spectrum (Figure \ref{fig:even-rank}), or be heavily skewed towards one end (Fig. \ref{fig:high-rank} and \ref{fig:low-rank}). In these common scenarios, traditional matchmaking metrics that rely on simple statistical measures of central tendency, such as the mean or median rank, prove inadequate. As demonstrated in Fig. \ref{fig:even-rank}-\ref{fig:middle-rank}, the mean or median can be a poor representative of the team as a whole. Matching based on these metrics can create a disastrous experience for players far from the average, forcing low-ranked players into overwhelmingly difficult matches or high-ranked players into trivial ones, ultimately frustrating all participants.

This paper proposes Cinder, a novel, two-stage matchmaking system designed to create fair matches even for lobbies with high skill variance. The system is built on two key principles:
\begin{enumerate}
    \item \textbf{Fast Preliminary Filtering:} A computationally inexpensive preliminary check to quickly discard incompatible lobby pairings.
    \item \textbf{Accurate Fairness Quantification:} A following precise check that considers the skill distribution of all players in the lobby to generate a single fairness score.
\end{enumerate}

This paper is structured as follows: Section \ref{sec:center_of_ratings} details the first stage of our system, a preliminary filter based on a non-outlier range overlap. Section \ref{sec:buckets} describes our non-linear skill bucketing method. Section \ref{sec:fairness} introduces our primary fairness metric, the ``Sanction Score," based on the Wasserstein distance. Section \ref{sec:results} presents the distribution of this score based on large-scale simulation. Finally, Section \ref{sec:conclusion} concludes the paper and suggests avenues for future work.

\section{Center of Ratings}
\label{sec:center_of_ratings}

\begin{figure}[h]
    \centering
    \includegraphics[width=0.8\textwidth]{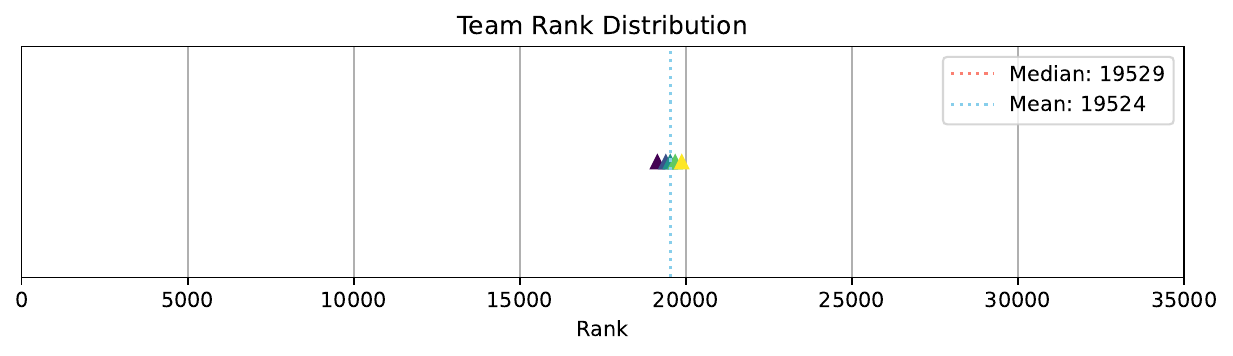}
    \caption{Equally Ranked Lobby}
    \label{fig:equal-rank}
\end{figure}

Before committing to a computationally expensive comparison of two lobbies, we employ a preliminary filter. The goal of this filter is to quickly identify and discard obviously poor matches. This filter is based on comparing the primary cluster of players in each lobby, which we define as the non-outlier range.

As popular rating systems such as ELO \cite{elo} or Glicko-2 \cite{glicko} often lack a theoretical upper limit for ratings, for practical purposes, we establish a mathematical cap, $x_{ucap}$, and a lower cap, $x_{lcap}$ (which often is $0$). Any rating above $x_{ucap}$ is treated as $x_{ucap}$, and any rating below $x_{lcap}$ is treated as $x_{lcap}$.

\begin{figure}[h]
    \centering
    \includegraphics[width=0.8\textwidth]{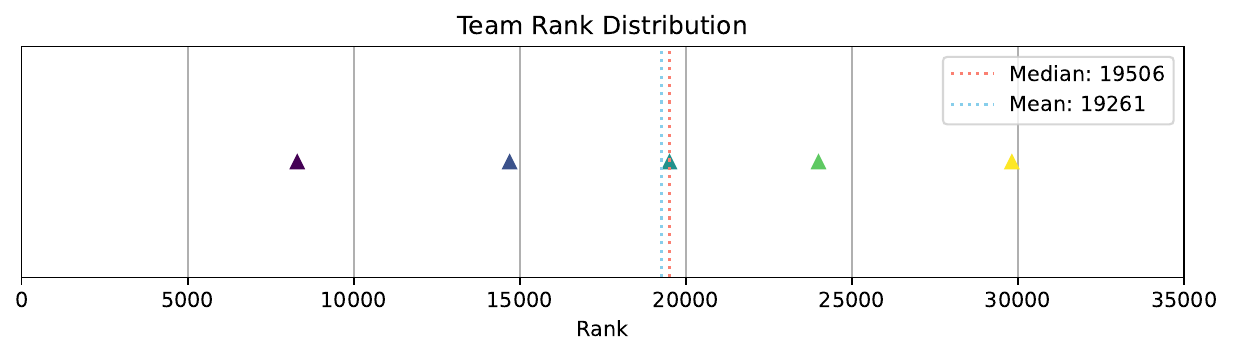}
    \caption{Evenly Distributed Ranked Lobby}
    \label{fig:even-rank}
\end{figure}

\begin{figure}[h]
    \centering
    \includegraphics[width=0.8\textwidth]{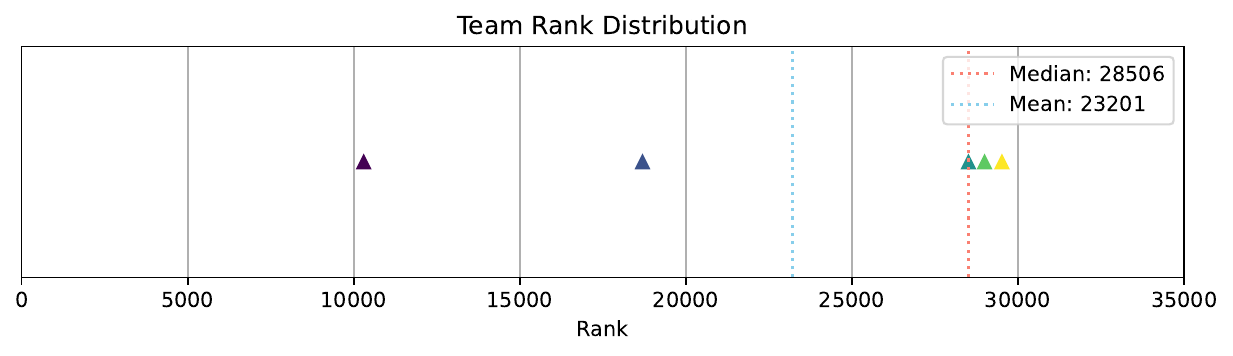}
    \caption{High Ranked Lobby}
    \label{fig:high-rank}
\end{figure}

\begin{figure}[h]
    \centering
    \includegraphics[width=0.8\textwidth]{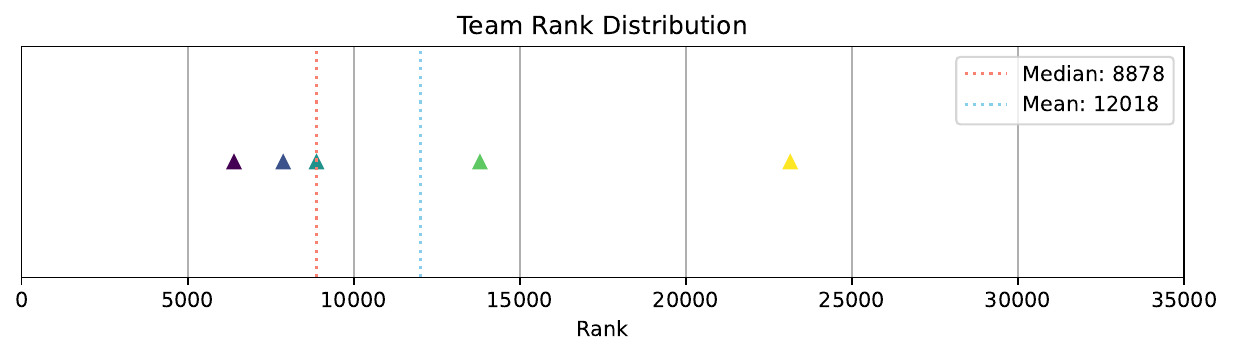}
    \caption{Low Ranked Lobby}
    \label{fig:low-rank}
\end{figure}

\begin{figure}[h]
    \centering
    \includegraphics[width=0.8\textwidth]{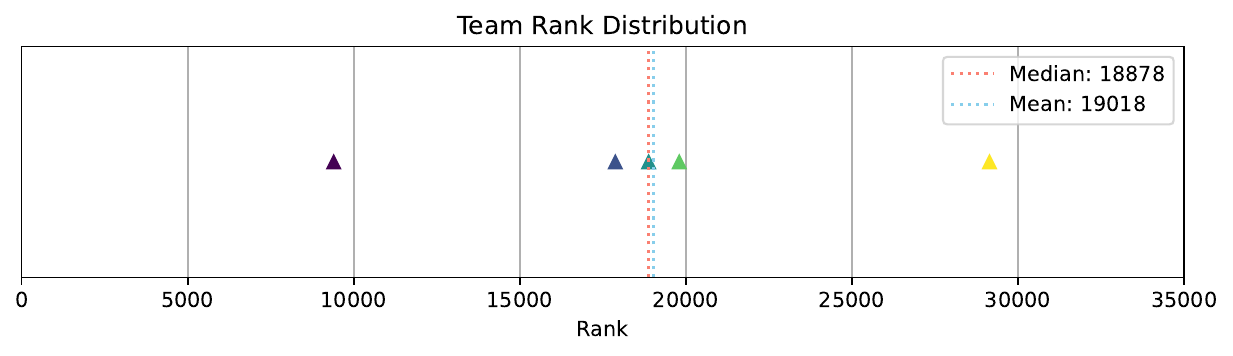}
    \caption{Medium Ranked Lobby}
    \label{fig:middle-rank}
\end{figure}

\begin{figure}[h]
    \centering
    \includegraphics[width=0.8\textwidth]{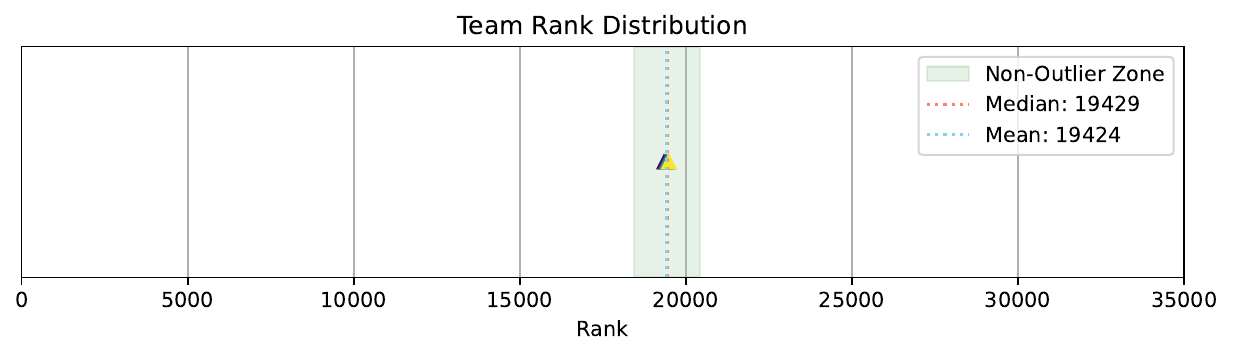}
    \caption{Equally Ranked Lobby with Non-Outlier Range}
    \label{fig:equal-rank-outlier}
\end{figure}

\begin{figure}[h]
    \centering
    \includegraphics[width=0.8\textwidth]{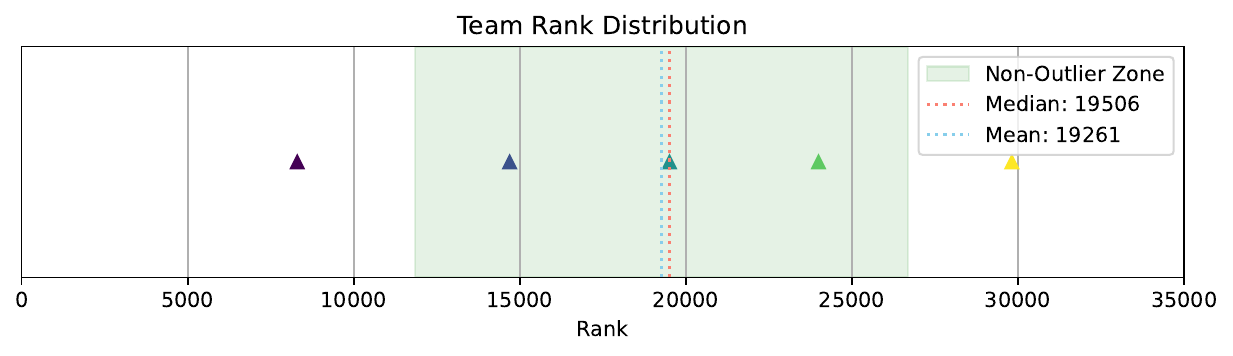}
    \caption{Evenly Distributed Ranked Lobby with Non-Outlier Range}
    \label{fig:even-rank-outlier}
\end{figure}

\begin{figure}[h]
    \centering
    \includegraphics[width=0.8\textwidth]{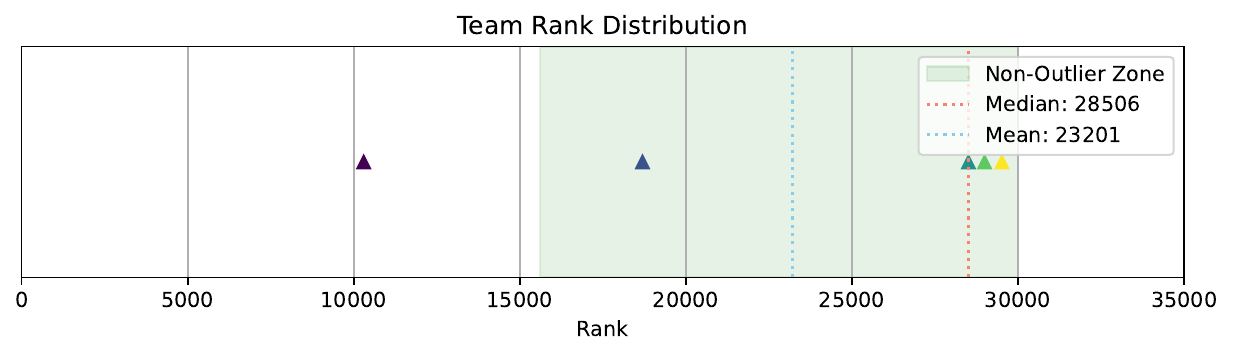}
    \caption{High Ranked Lobby with Non-Outlier Range}
    \label{fig:high-rank-outlier}
\end{figure}

\begin{figure}[h]
    \centering
    \includegraphics[width=0.8\textwidth]{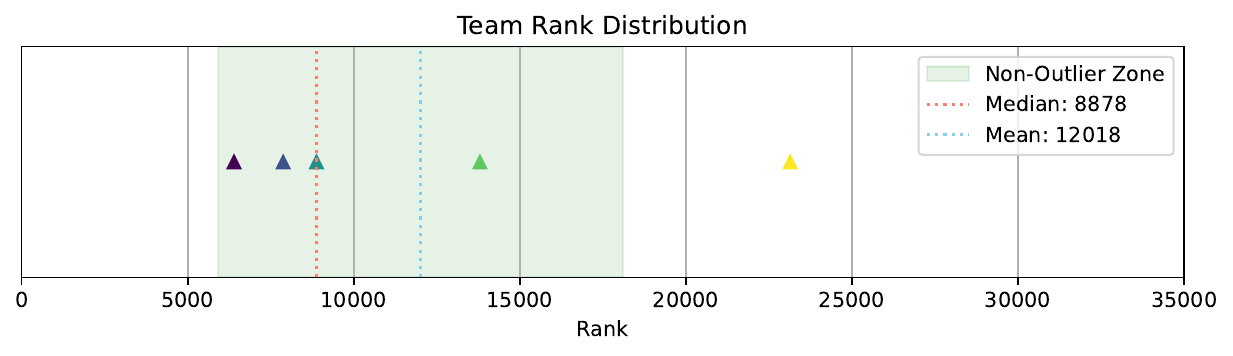}
    \caption{Low Ranked Lobby with Non-Outlier Range}
    \label{fig:low-rank-outlier}
\end{figure}

\begin{figure}[h]
    \centering
    \includegraphics[width=0.8\textwidth]{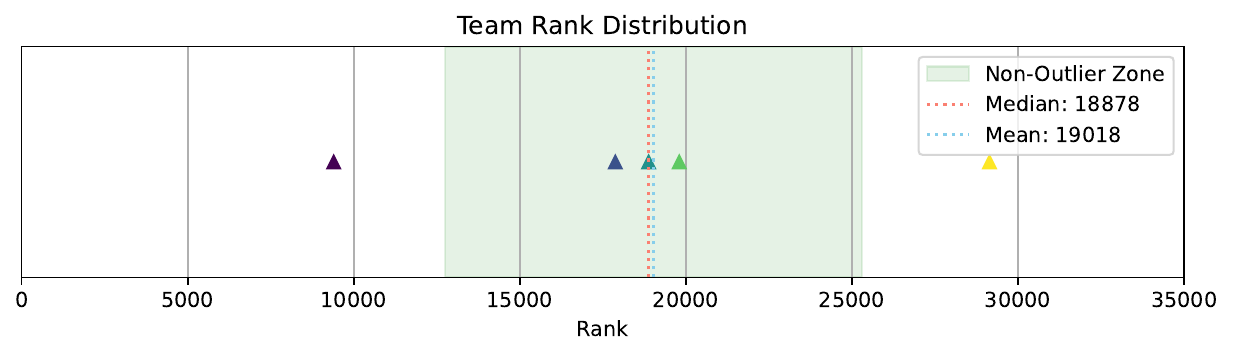}
    \caption{Medium Ranked Lobby with Non-Outlier Range}
    \label{fig:middle-rank-outlier}
\end{figure}

To find the non-outlier range, we use a modified Z-Score calculation. We first define a modified standard deviation, $\sigma'$, and then use it to define the lower and upper bounds of this range ($x_l$ and $x_u$), as shown in Eq. \ref{eq:z_score}. This modification ensures a minimum range width, preventing it from collapsing to zero in lobbies where all players have very similar ranks.

\begin{equation}
\label{eq:z_score}
\begin{split}
\sigma^{\prime} &= \max\{\sigma, \frac{x_{ucap} - x_{lcap}}{n_{\text{bucket}}}\} \\
x_{l} &= \mu - \sigma^{\prime} \\
x_{u} &= \mu + \sigma^{\prime}
\end{split}
\end{equation}

where $\mu$ and $\sigma$ are the mean and standard deviation of the lobby's ranks, and $n_{\text{bucket}}$ is the total number of skill buckets (detailed in Section \ref{sec:buckets}). Figures \ref{fig:equal-rank-outlier}-\ref{fig:middle-rank-outlier} illustrate the non-outlier ranges for the lobbies previously shown in Figures \ref{fig:equal-rank}-\ref{fig:middle-rank}.

\begin{align}
\label{eq:outlier}
\sigma' &= \max \{\sigma, (x_\text{ucap} - x_\text{lcap}) / n_\text{bucket} \} \nonumber \\
x_\text{l} &= \mu - \sigma' \nonumber \\
x_\text{u} &= \mu + \sigma'
\end{align}

\subsection{Range Overlap}

With a non-outlier range $[x_l, x_u]$ defined for each lobby, we can perform our preliminary compatibility test. We are interested in the overlap of these ranges between two lobbies (Lobby A and Lobby B). To quantify this overlap, we use the Ruzicka similarity index \cite{ruvzivcka} $S_{R}$, a continuous variant of the Jaccard similarity index, defined in Eq. \ref{eq:ruzicka}.

\begin{equation}
\label{eq:ruzicka}
S_{R}(A, B) = \frac{\sum_{i} \min(a_{i}, b_{i})}{\sum_{i} \max(a_{i}, b_{i})}
\end{equation}

\begin{figure}[h]
    \centering
    \includegraphics[width=0.8\textwidth]{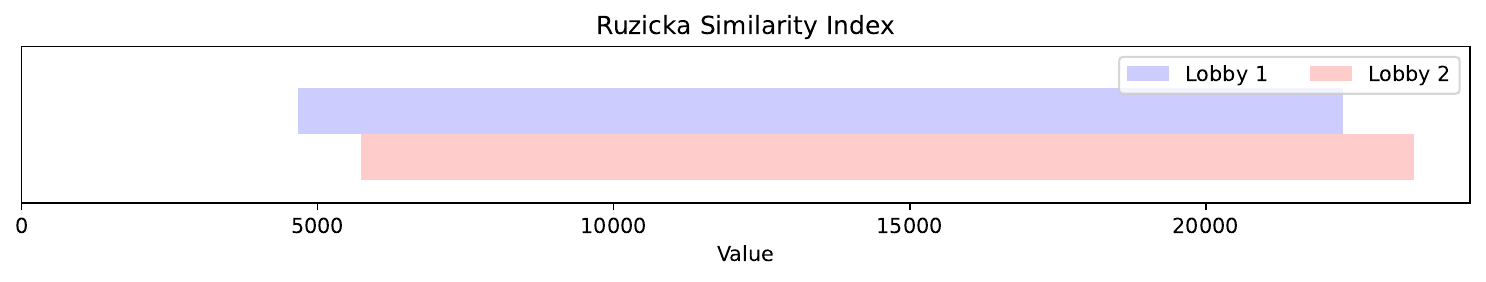}
    \caption{High Overlapping Lobbies}
    \label{fig:good-overlap}
\end{figure}

\begin{figure}[h]
    \centering
    \includegraphics[width=0.8\textwidth]{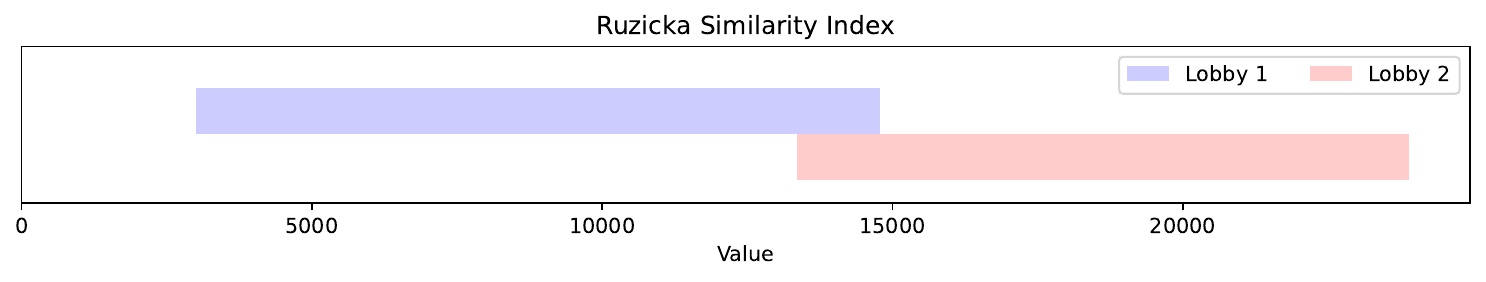}
    \caption{Low Overlapping Lobbies}
    \label{fig:bad-overlap}
\end{figure}

A high $S_{R}$ value indicates significant overlap between the core skill ranges of the two lobbies (Figure \ref{fig:good-overlap}), suggesting a potentially good match. A low $S_{R}$ value indicates a large disparity (Figure \ref{fig:bad-overlap}). Fig. \ref{fig:good-overlap} and \ref{fig:bad-overlap} have $S_R$ as $0.8796$ and $0.0684$ respectively. We can then define a minimum $S_{R}$ threshold to quickly reject unlikely lobby pairings, reducing the computational load on the more intensive second stage of the matchmaker.

\section{Buckets}
\label{sec:buckets}

Lobbies that pass the preliminary overlap filter proceed to a more detailed fairness evaluation. This evaluation is based on mapping the entire skill range ($x_{lcap}$ to $x_{ucap}$) into $n_{\text{bucket}}$ skill buckets.

A simple linear division (i.e., giving all buckets equal width) is suboptimal, as the majority of players are clustered around the median rank, as in the case of a normal distribution. This would result in poor skill differentiation for the largest segment of the player base. Instead, we require a non-linear distribution where buckets are narrower (providing higher granularity) near the center of the rank distribution and wider at the extremes.

\begin{figure}[h]
    \centering
    \includegraphics[width=0.8\textwidth]{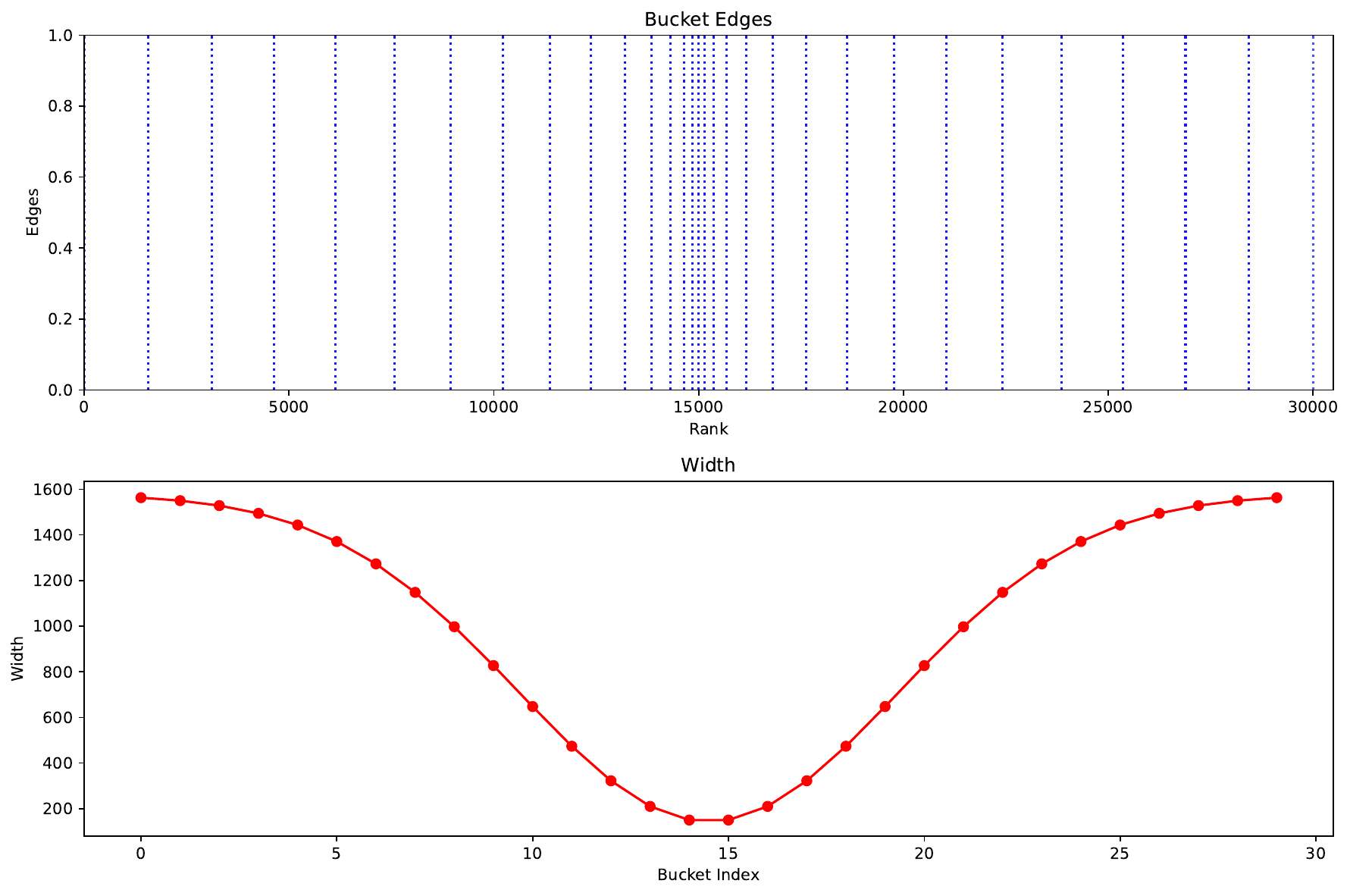}
    \caption{Rank Buckets}
    \label{fig:buckets}
\end{figure}

To achieve this, we use a modified inverted normal distribution to determine the width of each bucket, as shown in Figure \ref{fig:buckets}. We also enforce an empirical minimum bucket width, $W_{\text{min}}$ (e.g., 150 rank points), to ensure the central trough does not dip to zero. The final width $W_i$ of the $i$-th bucket is calculated using Eq. \ref{eq:bucket_width}. This method effectively scales the width according to the inverted normal PDF, concentrating precision where it is most needed.

\begin{equation}
\label{eq:bucket_width}
W_{i} = W_{\min} + (\phi_{\max} - \phi(x_{i})) \cdot \left[ \frac{R_{\text{total}} - (n_{\text{bucket}} \cdot W_{\min})}{\sum_{j=1}^{n_{\text{bucket}}} (\phi_{\max} - \phi(x_{j}))} \right]
\end{equation}

where:
\begin{itemize}
    \item $W_{\min}$ is the minimum allowed bucket width.
    \item $R_{\text{total}}$ is the total rank range ($x_{ucap} - x_{lcap}$).
    \item $\phi(x_{i})$ is the value of the normal probability density function (PDF) for the $i$-th bucket's position.
    \item $\phi_{\max}$ is the maximum value of the normal PDF (at the mean).
\end{itemize}

\section{Matchmaking Fairness}
\label{sec:fairness}

Once the skill buckets are defined, we can represent each player in a lobby by the index of the bucket their rating falls into. This discretization transforms the non-linear, continuous space of player ratings into a linear, discrete space of bucket indices.

This allows us to accurately quantify the dissimilarity between two lobbies using the Kantorovich distance \cite{kantorovich} or Wasserstein distance \cite{vaserstein}. As shown in Eq. \ref{eq:wasserstein}, the $W_1$ distance between two 1D distributions $U$ and $V$ of equal size $n$ is the sum of the absolute differences between their sorted elements.

\begin{equation}
\label{eq:wasserstein}
W_{1}(U, V) = \sum_{i=1}^{n} |u_{(i)} - v_{(i)}|
\end{equation}
where $u_{(i)}$ and $v_{(i)}$ are the $i$-th sorted bucket indices for the players in lobby $U$ and lobby $V$, respectively.

For example, the highly overlapping lobbies from Figure \ref{fig:good-overlap} yield a low Sanction Score of 2.8, indicating a good match. Conversely, the lobbies with low overlap from Figure \ref{fig:bad-overlap} yield a high Sanction Score of 11.2.

We define this distance as the ``Sanction Score." A larger score signifies greater dissimilarity between the two lobbies and, thus, a less fair match. This step is distinct from the Ruzicka similarity index \cite{ruvzivcka} discussed earlier because it considers all players in the lobby, including outliers, rather than just the central non-outlier range.

With this metric, a matchmaker has two options:

\begin{enumerate}
    \item Iterate over all waiting lobbies to find the pairing with the absolute minimum Sanction Score.
    \item Define a maximum acceptable Sanction Score threshold and pair the first match that fall below it.
\end{enumerate}

The latter approach is significantly faster and provides consistently good results, provided the threshold is chosen well.

\section{Results}
\label{sec:results}

\begin{figure}[h]
    \centering
    \includegraphics[width=0.8\textwidth]{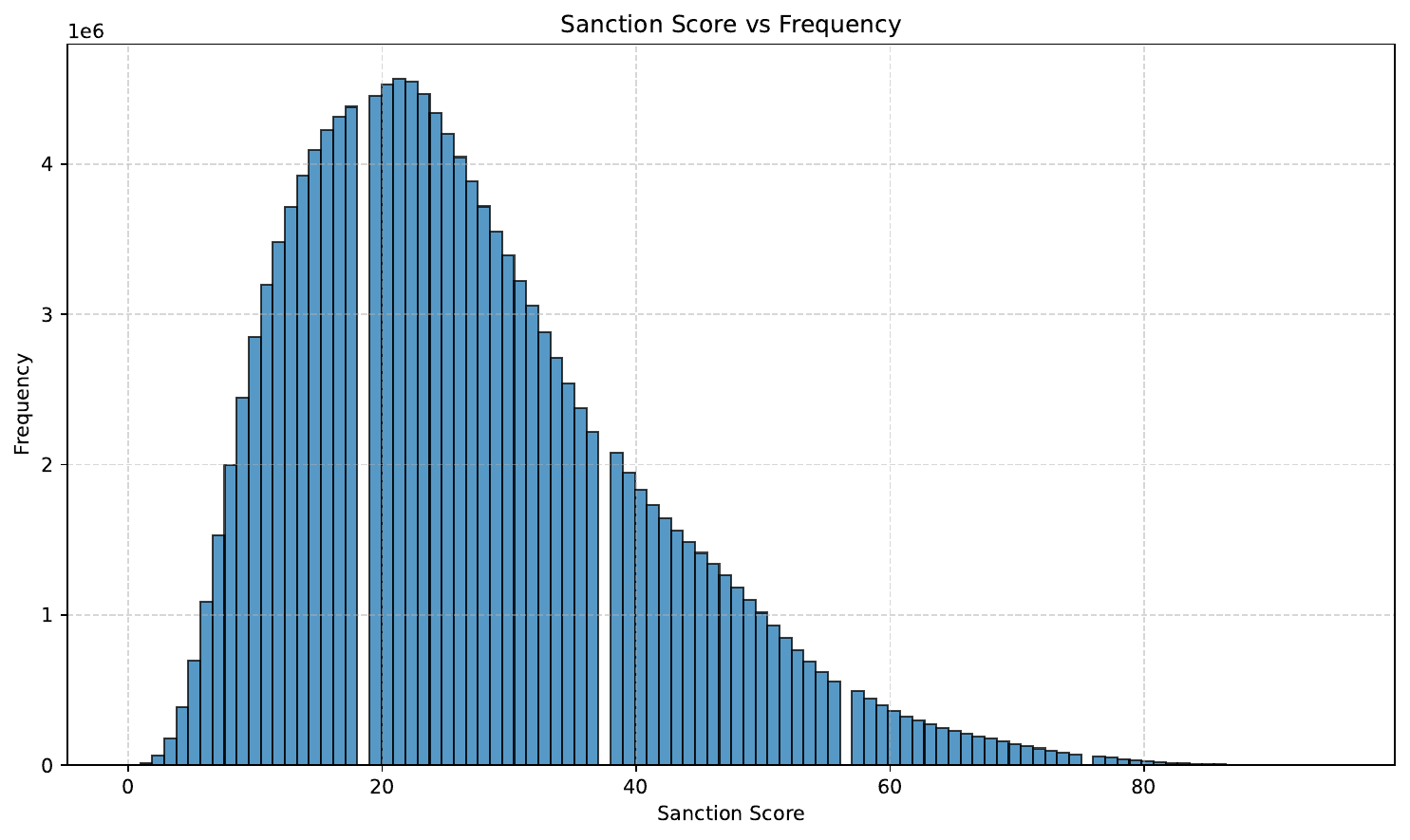}
    \caption{Sanction Score vs Frequency}
    \label{fig:sanction}
\end{figure}

To understand the practical behavior of the Sanction Score, we simulated 140,000,000 random lobby pairings. The distribution of the resulting Sanction Scores is shown in Figure \ref{fig:sanction}.

As can be seen in the histogram, the frequency of scores is not normally distributed; rather, it forms a right-skewed distribution. The distribution peaks at a Sanction Score of approximately 20-25 and has a long tail extending to the right. This indicates that while the vast majority of random pairings are clustered around a certain level of ``unfairness," extremely poor matches (high scores) are possible but increasingly rare. This empirical distribution is invaluable for setting an intelligent threshold for the matchmaking system, allowing for a data-driven trade-off between match quality and queue speed.

\section{Conclusion}
\label{sec:conclusion}

This paper introduced Cinder, a novel two-stage system for fast and fair matchmaking in online games. Cinder addresses the failures of popular matching by analyzing the entire skill distribution of lobbies. The system's first stage acts as a high-speed preliminary filter, using the Ruzicka similarity index \cite{ruvzivcka} to compare the non-outlier skill ranges of lobbies and discard incompatible pairings. The second, more precise stage quantifies match fairness by mapping all players to a non-linear, granularity-focused set of skill buckets. By calculating the 1D Wasserstein distance between the sorted bucket indices of two lobbies, we generate a ``Sanction Score" which is a single, robust metric that represents the dissimilarity, or unfairness, of a potential match between two lobbies.

Our analysis of 140 million simulated pairings demonstrates that this Sanction Score provides a predictable normal distribution, enabling the selection of an effective matchmaking threshold. This allows a system to balance match quality against queue time efficiently.

Future work could involve testing Cinder in a live environment to measure its impact on player satisfaction and queue dynamics. Further research could also explore the integration of other factors, such as player wait time or role preference, into the Sanction Score calculation, as well as optimizing the bucket distribution (Eq. \ref{eq:bucket_width}) based on a game's specific player-rank demographics.


\bibliographystyle{amsplain}
\bibliography{paper}

\end{document}